\documentclass[11pt,a4paper]{article}
\pdfoutput=1
\usepackage[table,xcdraw]{xcolor}
\usepackage[hyperref]{emnlp2020}
\usepackage{times}
\usepackage{latexsym}

\usepackage{mathtools}
\usepackage{booktabs}
\usepackage{graphicx}
\usepackage{hyperref}
\usepackage{listings}

\graphicspath{ {./figures/} }
\usepackage{microtype}

\usepackage{calrsfs}
\DeclareMathAlphabet{\pazocal}{OMS}{zplm}{m}{n}
\usepackage{bm}
\usepackage{bbm}
\usepackage{amssymb}
\usepackage{float}

\newenvironment{citemize}{\begin{list}{$\bullet$}{\topsep=\smallskipamount\itemsep=1pt\parsep=1pt\labelwidth=.5em}}{\end{list}}
\newenvironment{cenumerate}{\begin{list}{\labelenumi}{\usecounter{enumi}\topsep=.2\smallskipamount\itemsep=0pt\parsep=1pt\labelwidth=1.0em}}{\end{list}}

\aclfinalcopy 

\DeclareMathOperator*{\argmax}{arg\,max}
\DeclareMathOperator*{\geomean}{GM}

\newcommand{\secref}[1]{\hyperref[#1]{Section \ref*{#1}}}

\title{ÚFAL at MRP~2020:\ Permutation-invariant Semantic Parsing in PERIN}

\author{David Samuel \and Milan Straka \\
  Charles University \\
  Faculty of Mathematics and Physics \\
  Institute of Formal and Applied Linguistics \\
  \texttt{\{samuel,straka\}@ufal.mff.cuni.cz}\\}

\date{}

\begin{document}
\maketitle

\begin{abstract}
We present PERIN, a novel permutation-invariant approach to sentence-to-graph semantic parsing. PERIN is a versatile, cross-framework and language independent architecture for universal modeling of semantic structures. Our system participated in the CoNLL 2020 shared task, Cross-Framework Meaning Representation Parsing (MRP 2020), where it was evaluated on five different frameworks (AMR, DRG, EDS, PTG and UCCA) across four languages. PERIN was one of the winners of the shared task. The source code and pretrained models are available at \url{http://www.github.com/ufal/perin}.
\end{abstract}
\section{Introduction}

The aim of the CoNLL 2020 shared task, Cross-Framework Meaning Representation Parsing \cite[MRP 2020;][]{Oep:Abe:Abz:20}, is to translate plain text sentences into their corresponding graph-structured meaning representation.\hspace{-.08em}\footnote{See \url{http://mrp.nlpl.eu/2020/} for more details.} MRP 2020 features five formally and linguistically different frameworks with varying degrees of linguistic and structural complexity:

\begin{citemize}
  \item \textbf{AMR:} Abstract Meaning Representation \cite{banarescu2013abstract};
  \item \textbf{DRG:} Discourse Representation Graphs  \cite{abzianidze-etal-2017-parallel} provide a graph encoding of Discourse Representation Structure \cite{van1992presupposition};
  \item \textbf{EDS:} Elementary Dependency Structures \cite{oepen2006discriminant};
  \item \textbf{PTG:} Prague Tectogrammatical Graphs \cite{hajic2012announcing};
  \item \textbf{UCCA:} Universal Conceptual Cognitive Annotation \cite{abend2013universal}.
\end{citemize}

\goodbreak

\noindent
These frameworks constitute the \emph{cross-framework} track of MRP 2020, while the separate \emph{cross-lingual} track introduces one additional language for four out of the five frameworks: Chinese AMR \cite{li2016annotating}, German DRG, Czech PTG and German UCCA \cite{hershcovich-etal-2019-semeval}.

In agreement with the shared task objective to advance uniform meaning representation parsing across diverse semantic graph frameworks and languages, we propose a language and structure agnostic sentence-to-graph neural network architecture modeling semantic representations from input sequences.

The main characteristics of our approach are:

\begin{citemize}
  \item \textbf{Permutation-invariant model:} PERIN is, to our best
  knowledge, the first graph-based semantic parser that predicts all nodes at once in parallel and trains them with a permutation-invariant loss function. Semantic graphs are naturally \emph{orderless}, so constraining them to an artificial node ordering creates an unfounded restriction; furthermore, our approach is more expressive and more efficient than \emph{order-based} auto-regressive models.
  \item \textbf{Relative encoding:} We present a substantial improvement of relative encodings of node labels, which map anchored tokens onto label strings \cite{Str:Str:19}. Our novel formulation allows using a richer set of encoding rules.
  \item \textbf{Universal architecture:} Our work presents a general sentence-to-graph pipeline adaptable for specific frameworks only by adjusting pre-processing and post-processing steps.
\end{citemize}

\noindent
Our model was ranked among the two winning systems in both the \emph{cross-framework} and the \emph{cross-lingual} tracks of MRP 2020 and significantly advanced the accuracy of semantic parsing from the last year's MRP 2019.

\section{Related Work}

Examples of general, formalism-independent semantic parsers are scarce in the literature. \citet{hershcovich2018multitask} propose a universal transition-based parser for directed, acyclic graphs, capable of parsing multiple conceptually and formally different schemes. Furthermore, several participants of MRP 2019 presented universal parsers. \citet{Che:Dou:Xu:19} improved uniform transition-based parsing and used a different set of actions for each framework. \citet{Lai:Lo:Leu:19} submitted a transition-based parser with shared actions across treebanks, but failed to match the performance of the other parsers. \citet{Str:Str:19} presented a general graph-based parser, where the meaning representation graphs are created by repeatedly adding nodes and edges.

Graph-based parsers \cite{mcdonald2006online, peng2017addressing, dozat2018simpler, cai-lam-2020-amr} usually predict nodes in a sequential, auto-regressive manner and then connect them with a biaffine classifier. Unlike these approaches, our model infers all nodes in parallel while allowing the creation of rich intermediate representations by node-to-node self-attention.

Machine learning tools able to efficiently process unordered sets are gaining more attention in recent years. \citet{qi2017pointnet} and particularly \citet{zhang2019deep} proposed permutation-invariant neural networks for point clouds, which are of great relevance to our system. Our work was further inspired by \citet{carion2020end}, who utilize permutation invariance for object detection in a similar fashion to our sentence-to-graph generation.
\section{Methods}

\subsection{Graph Representation}
\label{sec:representation}

All five semantic formalisms share the same representation via directed labeled multigraphs in the graph interchange format proposed by \citet{kuhlmann2016towards}. Universally, the semantic units are represented by nodes and the semantic relationships by labeled edges. Each node can be anchored to a (possibly empty) set of input characters, and can contain a (possibly empty) list of properties, each being an attribute-value pair.

We simplify this graph structure by turning the properties into graph nodes: every property $\{\texttt{attribute}: \texttt{value}\}$ of node $n$ is removed and a new node with label $\texttt{value}$ is connected to the parent node $n$ by an edge labeled with $\texttt{attribute}$; the anchors of the new node are the same as of its parent.\hspace{-.08em}\footnote{``Nodeification'' of properties was motivated by the nature of AMR graphs, where the properties are equivalent to instanced concepts/nodes \cite{banarescu2013abstract}. From a~more practical viewpoint, it allows us to utilize a single classifier for both the node labels and the less-frequent properties, and to simplify the whole architecture.}  \autoref{fig:preprocess} illustrates this transformation together with other pre-processing steps (specific for each framework) explained in detail in \secref{sec:specifics}.

Another change to the internal graph representation is the use of relative label encoding (discussed in \secref{section:encoding}), which substitutes the original node labels by lists of relative encoding rules.

\subsection{Overall Architecture}
\label{sec:overall_architecture}

\begin{figure}[t]
\includegraphics[width=\columnwidth]{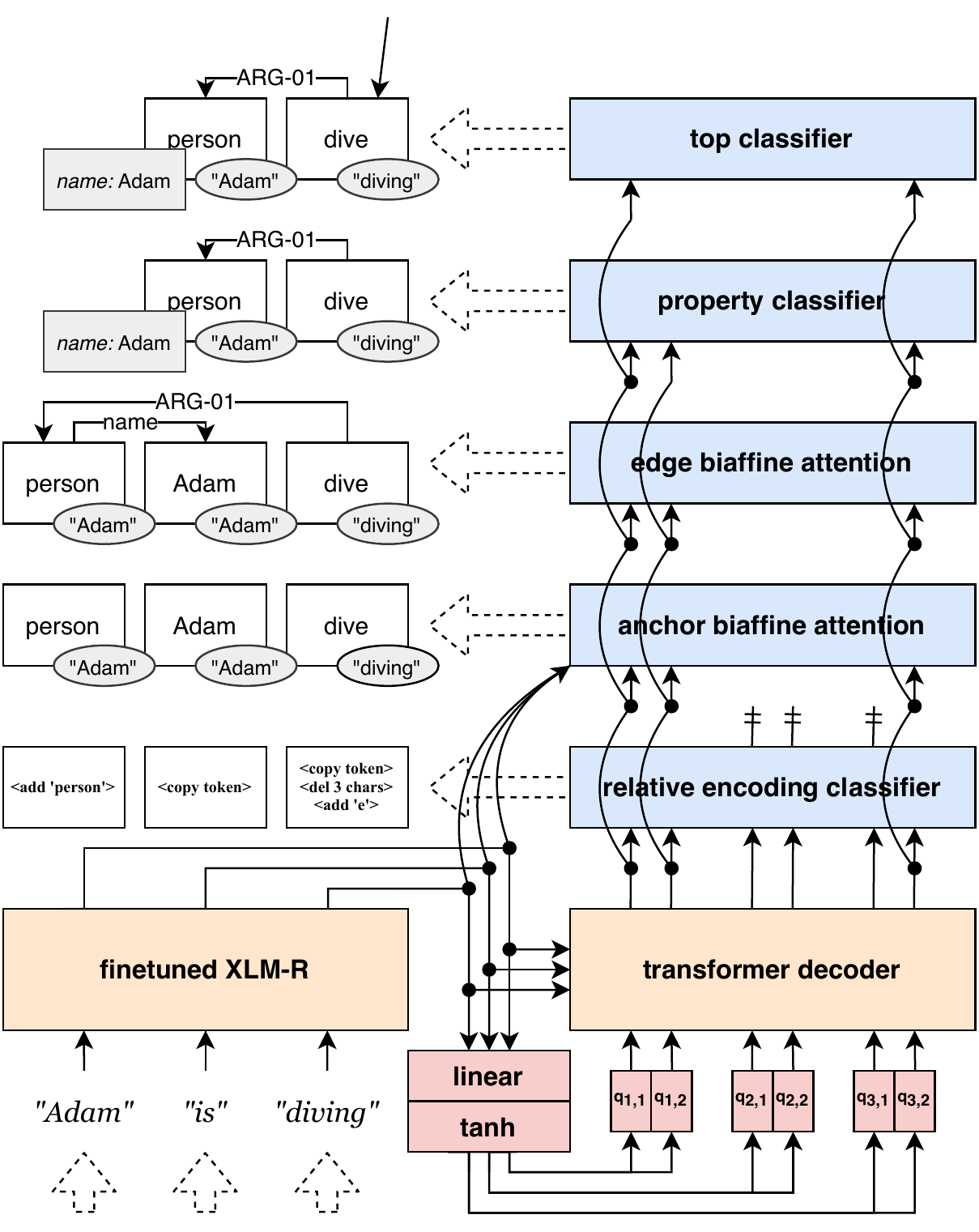}
\caption{Data flow through PERIN during inference. Every input token is processed by an encoder and transformed into multiple queries, which are further refined by a decoder. Each query is either \emph{denied} or \emph{accepted}, and the accepted ones are then gradually processed into the final semantic graph.}
\label{fig:inference}
\end{figure}

A simplified illustration of the whole model can be seen in \autoref{fig:inference}. The input is tokenized, an
encoder (\secref{section:xlm}) computes
contextual embeddings of the tokens, and each embedded token $\textbf{e}_i$ is then mapped onto $Q$ queries by nonlinear transformations $\textbf{q}_{i,t} = \textrm{tanh}\left(\textbf{W}_t\textbf{e}_i + \textbf{b}_t\right), t \in \{1,\dots Q\}$, where $\textbf{W}_t$ is a trainable weight matrix and $\textbf{b}_t$ is a trainable bias vector. After that, a decoder (Transformer with pre-norm residual connections \cite{nguyen2019transformers} and cross-attention into the contextual embeddings $\textbf{e}_i$) processes the queries, obtaining their final feature vectors $\textbf{h}_{i,t}$. These feature vectors are shared across all classification heads, each inferring specific aspects of the final meaning representation graph from them:

\begin{citemize}
  \item \textbf{Relative encoding classifier} decides what node label should serve as the ``answer'' to each query; a query can also be \emph{denied} (no node is created) when classified as ``null''. Relative label prediction is described in detail in \secref{section:mos}.
  
  \item \textbf{Anchor biaffine classifier} uses deep biaffine attention \cite{dozat2016deep} to create anchors between nodes and surface tokens -- to be more precise, the biaffine attention processes the latent vectors of queries $\textbf{h}_{i,t}$ and tokens $\textbf{e}_j$, and predicts the presence of an anchor between every pair of them as a binary classification task.
  
  \item \textbf{Edge biaffine classifier} uses three biaffine attention modules to predict whether an edge should exist between a pair of nodes (\emph{presence} binary classification), what label(s) should it have (\emph{label} multi-class or multi-label classification, depending on the framework) and what attribute should it have (\emph{attribute} multi-class classification) -- in essence, this module is a simple extension of the standard edge classifier by \citet{dozat2018simpler}.
  
  \item \textbf{Property classifier} uses a linear layer followed by a sigmoid nonlinearity to identify nodes that should be converted to properties.
  
  \item \textbf{Top classifier} uses a linear layer followed by a softmax nonlinearity (where the probabilities are normalized across nodes) to detect the \emph{top} node.
\end{citemize}

\noindent
This section described all modules capable of handling different characteristics of meaning representation graphs. Not all of them appear in each framework -- for example, AMR graphs do not need edge attributes, while UCCA graphs do not contain any properties. More details about specific framework configurations are given in \secref{sec:specifics}. 
\subsection{Permutation-invariant Graph Generation}

\begin{figure*}[t]
\centering
\includegraphics[width=0.7\textwidth]{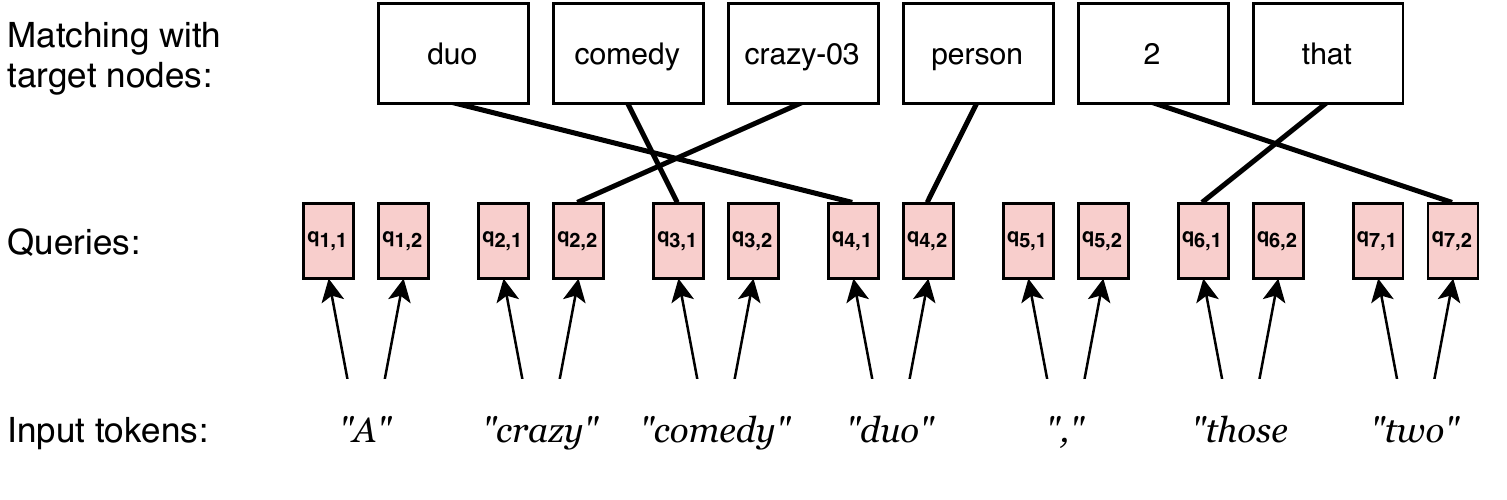}
\caption{Example of a matching between queries and target nodes during training. Every input token is mapped onto $Q$ (2 in this case) queries $\mathbf{q}_{i,j}$, which are decoded into node predictions $\hat{\mathbf{y}}_{i,j}$. These predictions are paired with the ground truth nodes $\mathbf{y}$, as in \autoref{eq:optimal_matching}. Then, the loss functions are computed with respect to the paired target nodes. Queries without any match should be classified as ``null'' nodes. When classified as ``null'' during inference, the query is not turned into any node (the query is \emph{denied}).}
\label{fig:matching}
\end{figure*}

Semantic graphs are \emph{orderless}, so it is unnatural to constrain their generation by some artificial node ordering. Traditionally, graph nodes have been predicted by a sequence-to-sequence model \cite{peng2017addressing}, with the nodes being generated in some hardwired order \cite{zhang2019amr}. Demanding a fixed node ordering causes the \emph{discontinuity issue} \cite{zhang2019deep}: even when correct items are predicted, they are viewed as completely wrong if not in the expected order. We avoid this issue by using such a loss function and such a model that produce the same outcome independently on the node ordering \cite{zaheer2017deep}. 

\subsubsection{Permutation-equivariant Model}

We transform the queries $\textbf{q} = \{\textbf{q}_i\}_{i=1}^N$ into hidden features $\textbf{h} = \{\textbf{h}_i\}_{i=1}^N$ in such manner that any permutation $\pi \in \mathfrak{G}_N$ of the input $\pi(\textbf{q}) = \{\textbf{q}_{\pi(i)}\}_{i=1}^N$ produces the same -- but permuted -- output
$\pi(\textbf{h}) = \{\textbf{h}_{\pi(i)}\}_{i=1}^N$. The Transformer architecture \cite{vaswani2017attention} conveniently fulfills this requirement (assuming positional embeddings are not used). Furthermore, it can combine any pair of input items independently of their distance and in an efficient non-autoregressive way.

\subsubsection{Permutation-invariant Loss}
\label{sec:pi_loss}

The hidden features $\textbf{h}$ are further refined into predictions $\hat{\textbf{y}} = f_{\bm{\theta}}(\mathbf{h})$ by the classification heads. In order to create a permutation-invariant loss function, i.e., a function $\pazocal{L}(\pi(\hat{\textbf{y}}), \textbf{y})$ giving the same result for every $\pi \in \mathfrak{G}_N$, we find a permutation $\pi^* \in \mathfrak{G}_N$ assigning each query to its most similar node. After permuting the targets according to $\pi^*$, the standard losses can be 
computed, because they are no longer dependent on the original ordering of $\hat{\textbf{y}}$ and $\textbf{y}$.\hspace{-.08em}\footnote{Unfortunately, the uniqueness of $\pi^*$ is not always guaranteed: given that the proposed matching depends only on labels and anchors, there might be multiple equivalent nodes (considering only labels and anchors). We break ties between such nodes by also minimizing the likelihood of their edges across all their permutations.}

To find the minimizing permutation $\pi^*$, we start by extending the (multi)set of target nodes $\textbf{y}$ by ``null'' nodes (denoted as $\varnothing$) in order to fulfill $\left\vert \hat{\textbf{y}} \right\vert = \left\vert \textbf{y} \right\vert$. When classified as ``null'' during inference, the query is \emph{denied} and omitted from further processing. The permutation $\pi^*$ is then defined as

\begin{equation} \label{eq:optimal_matching}
\pi^* = \argmax_{\pi \in \mathfrak{G}_N} \sum_{i=1}^{N}p_{\textrm{match}}(\hat{\textbf{y}}_i, \mathbf{y}_{\pi(i)}),
\end{equation}
where the matching score $p_{\textrm{match}}$ is composed of a~\emph{label score} and the geometric mean ($\geomean$) of \emph{anchor scores} of all input tokens $T$. The \emph{label score} of the $i^\textrm{th}$ query and the $j^\textrm{th}$ node is defined as the predicted probability of the target $j^\textrm{th}$ label; the \emph{anchor score} of the $i^\textrm{th}$ query, $j^\textrm{th}$ node and a~token $t \in T$ is defined as the predicted probability of the actual (non)existence of an anchor between $t$ and the $j^\textrm{th}$ node:
\begin{align*}
    p_{\textrm{match}} &= p_{\textrm{label}} \cdot \bar{p}_{\textrm{anchor}} \\
    p_{\textrm{label}}(\hat{\textbf{y}}_i, \mathbf{y}_j) &= \mathbbm{1}_{\mathbf{y}_j^{\textrm{label}} \neq \varnothing} P\big(\mathbf{y}_j^{\textrm{label}} \vert \mathbf{h}_i; {\bm{\theta}}\big) \\
    \bar{p}_{\textrm{anchor}}(\hat{\textbf{y}}_i, \mathbf{y}_j) &= \geomean_{t \in \textrm{tokens}} P\big(\mathbbm{1}_{t \in \textbf{y}_j^{\textrm{anchors}}} \vert t, \textbf{h}_i; \bm{\theta}\big).
\end{align*}
We use the geometric mean to keep the anchor score $\bar{p}_{\textrm{anchor}}$ magnitude independent of the number of tokens, and therefore have a similar weight as the label score $p_{\textrm{label}}$.

The optimal matching $\pi^*$ can be efficiently computed by the Hungarian algorithm \cite{kuhn1955hungarian} in $O(n^3)$. As a result, every query is assigned either to a regular node or to a ``null'' node $\varnothing$.  An illustration of a matching between queries and target nodes is presented in \autoref{fig:matching}.

The loss functions for the queries are computed with respect to the matched nodes. After finding $\pi^*$, we permute all target nodes and compute the classification losses in the standard ``order-based'' way (i.e., by minimizing the cross-entropy between the predictions and the corresponding targets). The losses of queries matched to the ``null'' nodes are ignored, except for their relative label loss $\ell_{\textrm{label}}$, which pushes these queries to predict $\varnothing$ as their label. The label loss is further altered by the focal loss factor \cite{lin2017focal} to mitigate the imbalance of labels introduced by extending the targets with the ``null'' nodes.

\subsubsection{Anchor Masking}
\label{sec:anchor_masking}
During the early experiments with this architecture, we noticed that nodes tend to be generated from their anchored tokens (or more precisely from the queries of their anchored tokens), after the outputs stabilize during first epochs. We employ this observation to create an inductive bias by limiting the possible pairings to occur only between target nodes and predictions from their anchored tokens. Formally, this is achieved by setting
$$\bar{p}_{\textrm{anchor}}\big(f_{\bm{\theta}}(\mathbf{h}_i), \mathbf{y}_j\big) = \varepsilon,$$
if the $j^\textrm{th}$ node is not anchored to the $i^\textrm{th}$ token, with $\varepsilon$ being a small positive constant close to $0$.
\subsection{Relative Label Encoding}
\label{section:encoding}

Similarly to \citet{straka2019udpipe,Str:Str:19}, we use \emph{relative encodings} for the prediction of node labels: instead of direct classification of label strings, we utilize rules specifying how to transform anchored surface tokens into the semantic labels. For example, in \autoref{fig:inference}, the anchored token \emph{``diving''} is transformed into \emph{``dive''} by using a relative encoding rule deleting its last three characters and appending a character \emph{``e''}. Such a rule could be also employed for predicting a node anchored to \emph{``taking''} or \emph{``giving''}. Relative encoding of labels is thus able to reduce the number of classification targets and generalize outside of the set of ``absolute'' label strings seen during training. Alternatively, the relative encoding can be seen as an extension of the pointer networks \cite{gu2016incorporating}, which also decides how to post-process the copied tokens. \autoref{tab:relative} demonstrates how the relative encoding rules reduce the number of targets that need to be classified.

\subsubsection{Minimal Encoding Rule Set}
\label{sec:min_encoding}

\begin{table}
\centering
\resizebox{\columnwidth}{!}{%
\begin{tabular}{@{}llrr@{}}
\toprule
\textbf{framework} & \textbf{language} & \textbf{\# labels strings} & \textbf{\# encoding rules} \\ \midrule
AMR & English & 27,049 & 4,385 \\
    & Chinese & 30,949 & 2,560 \\ \midrule[0.1pt]
DRG & English & 5,715  & 684   \\
    & German  & 1,905  & 918   \\ \midrule[0.1pt]
EDS & English & 31,933 & 1,322 \\ \midrule[0.1pt]
PTG & English & 39,336 & 529   \\
    & Czech   & 38,448 & 1,321 \\ \bottomrule
\end{tabular}%
}

\caption{The numbers of absolutely and relatively encoded node labels. Relative encodings lead to a significant reduction of classification targets in an order of magnitude across all frameworks. Note that node labels are the union of labels and property values (except for PTG), as described in \secref{sec:representation}.}
\label{tab:relative}
\end{table}

Naturally, a label can be generated from anchored tokens in multiple ways. Unlike previous works that needed some heuristic to select a single rule from all suitable ones \citep{Str:Str:19}, we do not constraint the space of the possible rules much. Instead, we construct the final set of encoding rules to be the smallest possible one capable of encoding all labels.

Formally, let $\pazocal{S}$ be an arbitrary class of functions transforming a list of text strings (anchored tokens) into another string (node label), and let $\mathsf{N}$ be the set of all nodes from the training set. For \hbox{$n \in \mathsf{N}$}, denote $n_t$ the anchored surface tokens and $n_\ell$ the target label string. Then the set of applicable rules for the node $n$ is $\pazocal{S}_n = \left\{r \in \pazocal{S} | r(n_t) = n_\ell \right\}$. Our goal is to find the smallest subclass $\pazocal{S}^* \subseteq \pazocal{S}$ capable of encoding all node labels, in other words a subclass $\pazocal{S}^*$ satisfying
$$
  \forall n \in \mathsf{N}: \pazocal{S}^* \cap \pazocal{S}_n \neq \emptyset.
$$
This formulation is equivalent to the minimal hitting set problem. Therefore, we can find the optimal solution of our problem by reducing it to a weighted MaxSAT formula in CNF: every $\pazocal{S}_n = \left\{r_1, r_2, \ldots, r_k \right\}$ becomes a hard clause $(r_1 \vee r_2 \vee \ldots \vee r_k)$ and every $r \in \pazocal{S}$ becomes a soft clause $(\neg r)$. 
We then submit this formula to the RC2 solver \cite{ignatiev2019rc2} to obtain the minimal set of rules. Note that although solving this problem can take up to several hours, it needs to be done only once and then cached for all the training runs.

\subsubsection{Space of Relative Rules}

Our space of relative rules $\pazocal{S}$ consists of four disjoint subclasses:

\begin{cenumerate}
    \item \emph{token rules} are represented by seven-tuples $(d_l, d_r, s, r_l, r_r, a_l, a_r)$ and process a list of anchored tokens $n_t$ by first deleting the first $d_l$ and the last $d_r$ tokens, then by concatenating the remaining ones into one text string with the separator $s$ inserted between them, followed by removing the first $r_l$ and last $r_r$ characters and finally by adding the prefix $a_l$ and suffix $a_r$;\footnote{To show a real example of a \emph{token rule} from EDS, the rule (\texttt{0}, \texttt{1}, \texttt{+}, \texttt{0}, \texttt{0}, \texttt{\char`_}, \texttt{\char`_a\char`_1}) maps tokens (``\texttt{at}'', ``\texttt{the}'', ``\texttt{very}'', ``\texttt{least}'', ``\texttt{,}'') into the label ``\texttt{\char`_at+the+very+least\char`_a\char`_1}''.}
    
    \item \emph{lemma rules} are created similarly to the token rules, but use the provided lemmas instead of tokens;
    
    \item \emph{number rules} transform word numerals into digits -- for example, tokens $[$ \emph{``forty''}, \emph{``two''}~$]$ become \emph{``42''};
    
    \item \emph{absolute rules} use the original label string $n_\ell$, without taking into account any anchored tokens $n_t$; they serve as the fallback rules when no relative encodings are applicable.
\end{cenumerate}


\subsubsection{Prediction of Relative Rules}
\label{section:mos}

Even with the minimal set of rules $\pazocal{S}^*$, multiple rules may be applicable to a single node. Therefore,
the prediction of relative rules is a multi-label classification problem. The target distribution for a node $n$ over all $r \in \pazocal{S}^*$ is defined as follows:\footnote{The target distribution is further modified by label smoothing \cite{szegedy2016rethinking} for better regularization.}
$$
    P(r \vert n) = 
\begin{dcases}
    \frac{1}{\left\vert \pazocal{S}^* \cap  \pazocal{S}_n \right\vert},& \text{if } r \in \pazocal{S}_n; \\
    0,              & \text{otherwise}.
\end{dcases}
$$
The label loss $\ell_{label}$ is then calculated as the cross-entropy between the target and the predicted distributions.

We use mixture of softmaxes (MoS) to mitigate the softmax bottleneck \cite{yang2017breaking} that arises when multiple hypotheses can be correctly applied to a single input. MoS allows the model to consider $K$ different hypotheses at the same time and weight them relatively to their plausibility.

Formally, let $\textbf{h}_q$ be the final latent vector for query $q$ and let $\textbf{W}_k$, $\textbf b_k$, $\textbf{w}_k$, $b_k$, $\textbf{w}_r$, $b_r$ be the trainable weights. Then, the estimated MoS distribution of relative rules $P_{\bm{\theta}}(r | n)$ is defined as follows:
\begin{align*}
    \textbf{x}_k &= \tanh(\textbf{W}_k\textbf{h}_q + \textbf b_k) \\
    \pi_k &= \frac{\operatorname{sigmoid}(\textbf{h}_q^\top \textbf{w}_k + b_k)}{\sum_{k'}{\operatorname{sigmoid}(\textbf{h}_q^\top \textbf{w}_{k'} + b_{k'})}} \\
    P_{\bm{\theta}}(r \vert n) &= \sum_{k=1}^{K}{\pi_k \operatorname{softmax}(\textbf{x}_k^\top \textbf{w}_r + b_r)}.
\end{align*}
\subsection{Finetuning XLM-R}
\label{section:xlm}

\begin{figure}
\includegraphics[width=\columnwidth]{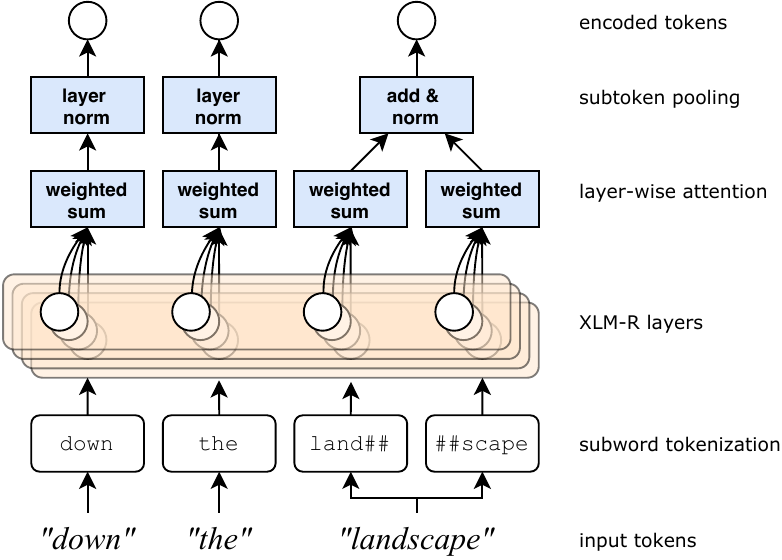}
\caption{Architecture of the encoder with finetuned XLM-R. The input tokens are first tokenized into subwords, which are then processed into contextual embeddings by layer-wise attention on the XLM-R intermediate layers. Finally, the subword embeddings are pooled to obtain the encoded tokens.}
\label{fig:encoder}
\end{figure}

To obtain rich contextual embeddings for each input token, we finetune the pretrained multilingual model XLM-R \cite{conneau2019unsupervised}.
The architecture of the encoder is presented in \autoref{fig:encoder}.

\subsubsection{Contextual Embedding Extraction}
\label{section:contextual_embedding_extraction}

Different layers in BERT-like models represent varying levels of syntactic and semantic knowledge \cite{van2019does}, raising a question of which layer (or layers) should be used to extract the embeddings from. Following \citet{kondratyuk201975}, we solve this problem by a purely data-driven approach and compute the weighted sum of all layers. Formally, let $\textbf{e}_l$ be the intermediate output from the $l^\textrm{th}$ layer and let $w_l$ be a trainable scalar weight. The final contextual embedding is then calculated as
$$\textbf{e} = \sum_{l=1}^{L}{\operatorname{softmax}(w_l)\textbf{e}_l}.$$
Note that each input token can be divided into multiple subwords by the XLM-R tokenizer. To obtain a single embedding for every token, we sum the embeddings of all its subwords. Finally, the contextual embeddings are normalized with layer normalization \cite{ba2016layer} to stabilize the training.\nobreak\hspace{-.08em}\footnote{A side effect of the normalization step is that the subword summation is equal to the more common subword average \cite{zhang2019amr}.}

\subsubsection{Finetuning Stabilization}

Given the large number of parameters in the pretrained XLM-R model, we employ several stabilization and regularization techniques in attempt to avoid overfitting.

We start by dividing the model parameters into two groups: the finetuned XLM-R and the rest of the network. Both groups are updated with AdamW optimizer \cite{loshchilov2017decoupled} 
, and their learning rate follows the inverse square root schedule with warmup \cite{vaswani2017attention}.
The learning rate of the finetuned encoder is frozen for the first 2000 steps before the warmup phase starts \cite{howard2018universal}. The warmup is set to 6000 steps for both groups, while the learning rate peak is $6 \cdot 10^{-5}$ for the XLM-R and $6 \cdot 10^{-4}$ for the rest of the network. The weight decay for XLM-R, $10^{-2}$, is considerably higher compared to $1.2 \cdot 10^{-6}$ used in the rest of the network \cite{devlin2018bert}.

Dropout of entire intermediate XLM-R layers results in additional regularization -- we drop each layer with 10\% probability by replacing $w_l$ with $-\infty$ during the final contextual embedding computation (\secref{section:contextual_embedding_extraction}). Inter-layer and attention dropout rates are the same as during the XLM-R pretraining.\hspace{-.08em}\footnote{Due to the space constrains, all hyperparameters for each training configuration (together with the source code and pretrained models) are published at \url{https://github.com/ufal/perin}.}
\subsection{Balanced Loss Weights}

\begin{figure*}[t]
\centering
\includegraphics[width=0.8\textwidth]{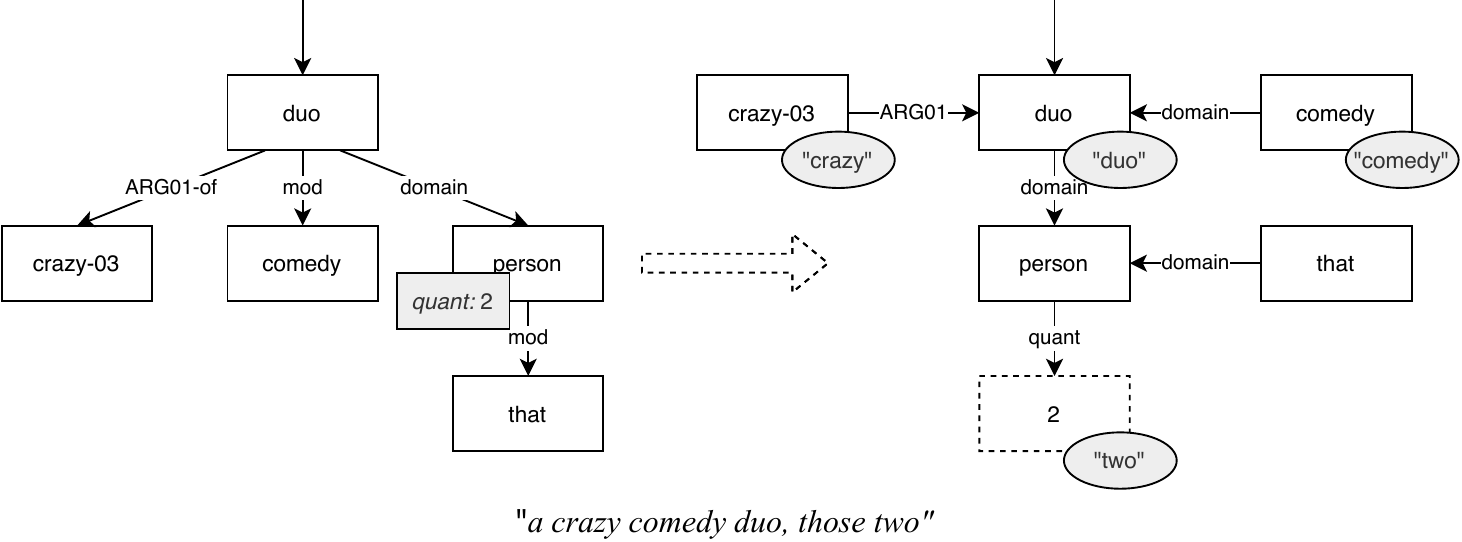}
\caption{Visualization of AMR pre-processing (\secref{sec:amr}) for the sentence \emph{``a crazy comedy duo, those two''}. The original graph is on the left and the transformed graph is shown on the right. Notice that the property \texttt{quant:2} of \texttt{person} is converted into a standalone node. The graph is normalized by reversing three inverted edges (note that \texttt{mod} is in fact \texttt{domain-of}) and some nodes get artificial anchors. Relative encoding rules are not included in this illustration for the sake of clarity, but it is worthwhile noting that nodes \texttt{person} and \texttt{that} contain only absolute label rules and are therefore not anchored.}
\label{fig:preprocess}
\end{figure*}

\begin{figure}
\includegraphics[width=\columnwidth]{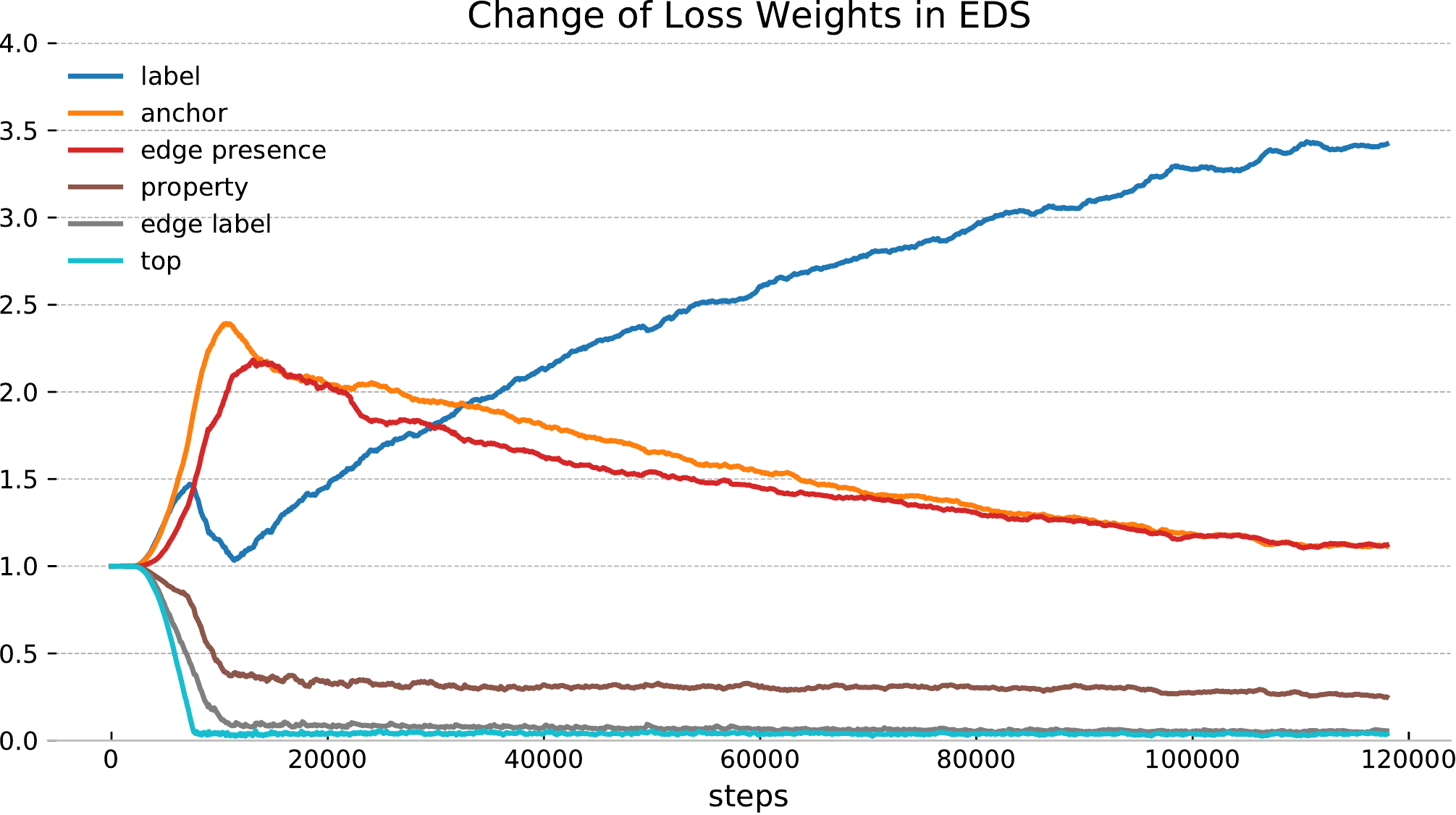}
\caption{Change of the loss weights throughout the training of an EDS parser. The relative difficulty of edge and anchor predictions seems to be higher at the beginning of the training, but then gradually decreases, allowing the model to concentrate primarily on label prediction.}
\label{fig:loss_weights}
\end{figure}

Semantic parsing is an instance of multi-task learning, where each task $t \in \mathsf{T}$ can have conflicting needs and where the task losses $\ell_t$ can have different magnitudes. The overall loss function $\pazocal{L}$ to be optimized therefore consists of the weighted sum of partial losses $\ell_t$:

$$\pazocal{L}(f_{\bm{\theta}}(\mathbf{x}), y) = \sum_{t \in \mathsf{T}}{w_t \ell_t(f_{\bm{\theta}}(\mathbf{x}), y)}.$$

\noindent
Finding optimal values for the loss weights $w_t$ is extremely complicated. This issue is usually resolved either by (suboptimally) setting all weights equally to 1 or by a thorough grid search. However, the complexity of the grid search grows exponentially with $\left\vert \mathsf{T} \right\vert$ and would need to be performed independently for all nine combinations of frameworks and languages.

A more feasible solution is to set the weights adaptively according to a data-driven metric as in \citet{kendall2018multi}. We follow \citet{chen2018gradnorm}, who balance the magnitudes of gradients $\left\Vert \nabla_{\bm{\theta_s}} w_i\ell_i \right\Vert_2$, where $\bm{\theta_s}$ are the weights of the shared part of the network. That magnitude is made proportional to the ratio of the current loss and its initial value: when $\ell_i$ decreases relatively quickly, its strength gets reduced to leave more space for the other tasks. Consequently, the loss weights $w_t$ are not static, but change throughout the training to balance the individual gradient norms. \autoref{fig:loss_weights} shows an example of the balancing dynamics.

\begin{table*}[t]
\small
\centering
\def\-{\rule[.5ex]{2em}{0.5pt}\kern.5em}
\def\={\rule[.5ex]{2em}{1pt}\kern.5em}
\catcode`@ = 13\def@{\bfseries}
\begin{tabular}{@{}lrrrrrr@{}}
\toprule
\textbf{System} & \textbf{AMR\textsuperscript{eng}} & \textbf{DRG\textsuperscript{eng}} & \textbf{EDS\textsuperscript{eng}} & \textbf{PTG\textsuperscript{eng}} & \textbf{UCCA\textsuperscript{eng}} & \textbf{Average} \\ \midrule
HUJI-KU \cite{Arv:Cui:Her:20} & 52.36\% & 62.75\% & 79.68\% & 53.76\% & 72.91\% & 64.29\% \\
Hitachi \cite{Oza:Mor:Kor:20} &@81.54\% & 93.19\% &@93.56\% & 88.73\% & 75.07\% & 86.42\% \\
HIT-SCIR \cite{Dou:Fen:Ji:20} & 69.80\% & 89.07\% & 87.40\% & 84.26\% & 74.76\% & 81.06\% \\
@ÚFAL PERIN   & 80.23\% &@94.16\% & 92.73\% & 88.44\% &@76.40\% & 86.39\% \\
@ÚFAL PERIN*  & 80.23\% &@94.16\% & 92.73\% &@89.19\% &@76.40\% &@86.54\% \\ \midrule
 &  &  &  &  &  &  \\ \midrule[1pt]
\textbf{System} & \textbf{AMR\textsuperscript{zho}} & \textbf{DRG\textsuperscript{deu}} & \= & \textbf{PTG\textsuperscript{ces}} & \textbf{UCCA\textsuperscript{deu}} & \textbf{Average} \\ \midrule
HUJI-KU \cite{Arv:Cui:Her:20} & 44.92\% & 62.33\% & \- & 58.49\% & 74.72\% & 60.11\% \\
Hitachi \cite{Oza:Mor:Kor:20} & 80.44\% &@93.36\% & \- & 87.35\% & 79.04\% & 85.05\% \\
HIT-SCIR \cite{Dou:Fen:Ji:20} & 49.39\% & 68.31\% & \- & 77.93\% & 80.02\% & 68.91\% \\
@ÚFAL PERIN   & 78.17\% & 89.83\% & \- & 91.27\% &@81.01\% & 85.07\% \\
@ÚFAL PERIN*  &@80.52\% & 89.83\% & \- &@92.24\% &@81.01\% &@85.90\% \\ \bottomrule
\end{tabular}
\caption{The \textbf{all} F1 scores and a macro-average total score of the shared task systems. PERIN is our official shared task submission and PERIN* is a post-competition submission with a fixed bug. The best results are typeset in \textbf{bold}. The top table contains the  \emph{cross-framework} scores on English treebanks, while the bottom table presents the \emph{cross-lingual} ones.}
\label{tab:framework_result}
\end{table*}

\subsection{Framework Specifics}
\label{sec:specifics}

\subsubsection{AMR}
\label{sec:amr}

AMR is a \emph{Flavor 2} framework, which means its nodes are not anchored to the surface forms. We instead exploit the general algorithm for the minimal encoding rule set (\secref{sec:min_encoding}) to create artificial anchors: considering all possible one-to-one anchors $a \in A_n$ for each node $n$, we infer all compatible rules $\pazocal{S}_n = \bigcup_{a \in A_n} \pazocal{S}_n^a$, and find the minimal set of rules $\pazocal{S}^*$. The artificial anchors of a node $n$ are then defined as $\{a \in A_n \vert \pazocal{S}_n^a \cap \pazocal{S}^* \neq \emptyset \}$. Consequently, our parser does not need any approximate anchoring (because we instead compute an anchoring minimizing the number of relative rules).

On the other hand, Chinese AMR graphs contain anchors (they are actually of \hbox{\emph{Flavor 1}}), therefore, the described procedure is applied only on English AMR.

AMR graphs also contains inverted edges that transform them into tree-like structures. The inverted edges are marked by modified edge labels (for example, \texttt{ARG0} becomes \texttt{ARG0-of}). We normalize the graphs back into their original non-inverted form, making them more uniform, simplifying edge prediction to become more local and independent of the global graph structure. An example of AMR pre-processing is shown in \autoref{fig:preprocess}.

Considering the fact that every node is artificially anchored to at most a single token, the anchor classifier is not needed, if anchor masking is used (\secref{sec:anchor_masking}). Finally, AMR parsing does not employ the edge attribute classifier.

\subsubsection{DRG}
Since the DRG graphs are also of \emph{Flavor 2}, they are pre-processed similarly to English AMR. Additionally, we reduce all nodes representing binary relations into labeled edges between the corresponding discourse elements.

Nodes in German DRG graphs are labeled in English, which decreases the applicability of relative encoding. Therefore, we employ the \texttt{opus-mt-de-en} \cite{tiedemann2020building} machine translation model from Huggingface's transformers package \cite{wolf2019huggingface} to translate the provided lemmas from German to English, before computing the relative encoding rules.

DRG parsing does not make use of anchor and edge attribute classifiers, just like AMR parsing.

\subsubsection{EDS}
EDS graphs are post-processed to contain a single continuous anchor for every node. The EDS parser contains all the classification modules described in \secref{sec:overall_architecture}, except for the edge attribute classifier.

\subsubsection{PTG}
\label{section:ptg}
Properties in PTG graphs are not converted into nodes as in other frameworks, but are directly predicted from latent vectors $\mathbf{h}_q$ by multi-class classifiers (one for each property type). Additionally, the \texttt{frame} properties are selected only from frames listed in CzEngVallex \cite{uresova2015czengvallex}.

We utilize all classification heads except for the top node classification, because PTG graphs contain special \texttt{<TOP>} nodes, which make the separate top prediction redundant.

\subsubsection{UCCA}

We augment the UCCA nodes by assigning them \texttt{leaf} and \texttt{inner} labels. Additionally, the \texttt{inner} nodes are anchored to the union of anchors of their children. Therefore, the nodes can be differentiated by the permutation-invariant loss (\secref{sec:pi_loss}).

The UCCA parser does not have the property classifier and the top classifier, where the latter is not needed, because the top node can be inferred from the structure of the rooted UCCA graphs.

\section{Results}

\begin{table*}[t]
\small
\centering
\def\-{\rule[.5ex]{2em}{0.5pt}\kern1em}
\catcode`@ = 13\def@{\bfseries}
\begin{tabular}{@{}lrrrrrrr@{}}
\toprule
\textbf{System} & \textbf{Tops} & \textbf{Labels} & \textbf{Properties} & \textbf{Anchors} & \textbf{Edges} & \textbf{Attributes} & \textbf{Average} \\ \midrule
HUJI-KU \cite{Arv:Cui:Her:20} & 85.85\% & 22.80\% & 29.48\% & 46.79\% & 61.35\% &  7.67\% & 62.43\% \\
Hitachi \cite{Oza:Mor:Kor:20} &@95.67\% & 68.93\% & 48.89\% & 61.95\% &@80.14\% & 24.93\% & 85.81\% \\
HIT-SCIR \cite{Dou:Fen:Ji:20} & 94.37\% & 61.84\% & 30.80\% & 52.18\% & 71.41\% & 22.51\% & 75.66\% \\
@ÚFAL PERIN*  & 94.20\% &@70.36\% &@49.34\% &@63.45\% & 79.68\% &@27.07\% &@86.26\% \\ \bottomrule
\end{tabular}%
\caption{Overall results for different MRP metrics, macro-averaged over all frameworks and languages. The best results are typeset in \textbf{bold}.}
\label{tab:score_result}
\end{table*}

\begin{table*}[t!]
\centering
\catcode`@ = 13\def@{\bfseries}
\small
\begin{tabular}{@{}lrrrrrr@{}}
\toprule
\textbf{Configuration}     & \textbf{Tops}    & \textbf{Labels}  & \textbf{Properties} & \textbf{Anchors} & \textbf{Edges}   & \textbf{Average} \\ \midrule
@ÚFAL PERIN*             & 89.53\%  & 93.45\%  & 94.34\%  & 93.40\%  & 90.74\%  & 92.73\% \\
w/o MoS             & 88.04\%  & 93.39\%  & 93.79\%  & 93.48\%  & 90.76\%  & 92.65\% \\
w/o focal loss      & 89.08\%  & 93.33\%  & 93.59\%  & 93.21\%  & 90.46\%  & 92.46\% \\
BERT encoder        & 89.95\%  & 92.97\%  & 94.74\%  & 92.92\%  & 89.84\%  & 92.27\% \\
w/o balanced losses & 89.23\%  & 92.28\%  & 94.46\%  & 92.12\%  & 89.19\%  & 91.60\% \\
\bottomrule
\end{tabular}%
\caption{Ablation study showing MRP scores of different configurations on EDS. The top row contains the submitted configuration without any changes; then we report the results for 1) label classifier without the mixture of softmaxes (MoS); 2) label loss not multiplied by the focal loss coefficient; 3) encoder with finetuned BERT-large (English) instead of multilingual XLM-R and 4) constant loss weights, equally set to $1.0$.}
\label{tab:ablation}
\end{table*}

\begin{table}[t]
\small
\centering
\catcode`@ = 13\def@{\bfseries}
\begin{tabular}{@{}lrrr@{}}
\toprule
@System          & @AMR\textsuperscript{eng} & @EDS\textsuperscript{eng} & @UCCA\textsuperscript{eng} \\ \midrule
best from MRP 2019 &  73.11\% &  92.55\% &  82.61\% \\
@ÚFAL PERIN*          & @78.43\% & @95.17\% & @82.71\% \\ \bottomrule
\end{tabular}
\caption{The last year's shared task had three frameworks -- English AMR, EDS and UCCA -- in common with MRP 2020. All parsers were evaluated on \emph{The Little Prince} dataset, the first row shows the F1 scores of the best performing parser for each framework \cite{Oep:Abe:Haj:19}.}
\label{tab:prince}
\end{table}

We present the overall results of our system in \autoref{tab:framework_result} and \autoref{tab:score_result}. Both tables contain F1 scores obtained using the official MRP metric.\hspace{-.08em}\footnote{Fine-grained results for each framework are available in the task overview by \citet{Oep:Abe:Abz:20}.} \autoref{tab:framework_result} shows the \textbf{all} F1 scores for the individual frameworks together with the overall averages for the \emph{cross-framework} and \emph{cross-lingual} tracks. Macro-averaged results (across all nine frameworks) for the different MRP metrics are displayed in \autoref{tab:score_result}.

Note that our original submission (denoted as PERIN) contained a bug in anchor prediction for Chinese AMR and both PTG frameworks. The bug caused the nodes to get anchored to at least one token. We submitted a fixed version called PERIN* in the post-competition evaluation and compare it with the original one in \autoref{tab:framework_result}.

According to the official whole-percent-only \textbf{all} F1 score, our competition submission reached tied first place in both the \emph{cross-lingual} and the \emph{cross-framework} track, with its performance virtually identical to the system by Hitachi~\cite{Oza:Mor:Kor:20}. Our bugfixed submission reached the first rank in both tracks, improving the \emph{cross-lingual} score by nearly one percent point. Our system excels in label prediction, which might suggest the effectiveness of the relative label encoding. Furthermore, our system surpasses the best systems from the last year's semantic shared task, MRP 2019 \cite{Oep:Abe:Haj:19}, by a wide margin -- as can be seen in \autoref{tab:prince}. 

PERIN falls short in AMR\textsuperscript{eng} parsing by 1.31~\%. On closer inspection, this follows from the inferior edge accuracy on this framework -- the difference to Hitachi is 4.56~\% on AMR\textsuperscript{eng} and 2.78~\% on AMR\textsuperscript{zho}. Furthermore, Hitachi is better in all aspects of EDS\textsuperscript{eng} and DRG\textsuperscript{deu}. On the other hand PERIN consistently beats Hitachi in both PTG and both UCCA frameworks. We hope that combining the strengths of these two parsers will help to further advance the state of meaning representation parsing.

\goodbreak
\subsection{Ablation Experiments}

We conducted several additional experiments to evaluate the effects of various components of our  architecture. The results are summarized in \autoref{tab:ablation}. We have decided to use EDS for these experiments because -- in our eyes -- it represents the ``average'' framework without any significant irregularities.

The experiments show that using the mixture of softmaxes for label prediction does not have a substantial effect and can be potentially omitted to reduce the parameter count. On the other hand, the inferior results of the model with constant equal loss weights demonstrate the importance of balancing them.
\section{Conclusion}

We introduced a novel permutation-invariant sentence-to-graph semantic parser called PERIN. Given its state-of-the-art performance across a number of frameworks, we believe permutation-invariant node prediction might be the first step in a promising direction of semantic parsing and generally of graph generation.
\vspace*{\fill}
\section*{Acknowledgments}

This work has been supported by the Grant Agency of the Czech Republic,
project EXPRO LUSyD (GX20-16819X). It has also been supported by the Ministry of Education, Youth and Sports of the Czech Republic, Project No. LM2018101 LINDAT/CLARIAH-CZ.

\clearpage

\bibliographystyle{acl_natbib}
\bibliography{mrp,perin}

\end{document}